\pdfoutput = 1
\documentclass{bmvc2k}

\usepackage{url}            % simple URL typesetting
\usepackage{booktabs}       % professional-quality tables
\usepackage{amsfonts}       % blackboard math symbols
\usepackage{nicefrac}       % compact symbols for 1/2, etc.
\usepackage{microtype}      % microtypography

\usepackage{amsmath,amssymb} % define this before the line numbering.
\usepackage{mathtools}
\usepackage{subfigure}

\usepackage{comment}
\usepackage{multirow}
\usepackage{multicol}
\usepackage{bbm}
\usepackage{color}

\usepackage{algorithm}
\usepackage{algorithmic}

\title{Supervised Incremental Hashing}

\addauthor{Bahadir Ozdemir}{ozdemir@cs.umd.edu}{1}
\addauthor{Mahyar Najibi}{najibi@cs.umd.edu}{1}
\addauthor{Larry S. Davis}{lsd@umiacs.umd.edu}{2}

\addinstitution{
Department of Computer Science\\
  University of Maryland \\
  College Park, MD 20742
}
\addinstitution{
  Institute for Advanced Computer Studies\\
  University of Maryland \\
  College Park, MD 20742
}

\runninghead{Ozdemir, Najibi, Davis}{Supervised Incremental Hashing}

\DeclareMathOperator*{\argmin}{\arg\!\:\min}
\begin{document}

\maketitle

\begin{abstract} 
 
We propose an incremental strategy for learning hash functions with kernels for large-scale image search. Our method is based on a two-stage classification framework that treats binary codes as intermediate variables between the feature space and the semantic space. In the first stage of classification, binary codes are considered as class labels by a set of binary SVMs; each corresponds to one bit. In the second stage, binary codes become the input space of a multi-class SVM. Hash functions are learned by an efficient algorithm where the NP-hard problem of finding optimal binary codes is solved via cyclic coordinate descent and SVMs are trained in a parallelized incremental manner. For modifications like adding images from a previously unseen class, we describe an incremental procedure for effective and efficient updates to the previous hash functions. Experiments on three large-scale image datasets demonstrate the effectiveness of the proposed hashing method, Supervised Incremental Hashing (SIH), over the state-of-the-art supervised hashing methods.
 
\end{abstract}

\section{Introduction}

Online image databases like Flickr grow steadily every day. Huge amounts of information lead to the requirement of efficient search algorithms. The linear time complexity of exhaustive similarity search can be reduced to sublinear time complexity for approximate nearest neighbor search through discretization \cite{Gionis:1999wf}. Several approaches have been proposed to discretize the feature space by preserving pairwise similarities between data points. Binary hashing methods, which map data points to short binary codes, have attracted attention from researchers in different areas due to their effectiveness in large-scale search and efficiency in memory storage. The methods for learning hash functions are categorized into unsupervised and supervised learning schemes. In this paper, we only focus on supervised hashing.

Supervised hashing methods are constructed to capture semantic similarity between images. Semantic information such as class labels or tags is used to regularize the process of learning binary codes. Several supervised methods have been proposed as extensions to unsupervised techniques by making them aware of the semantics of data. Iterative Quantization (ITQ) is used for both unsupervised and supervised hashing in conjunction with Principal Component Analysis (PCA) and Canonical Correlation Analysis (CCA), respectively \cite{Gong:2013kp}. A supervised version of BRE was proposed by Kulis and Darrell \cite{kulis2009learning}. Supervised Hashing with Kernels (KSH) \cite{Liu:2012ii} was introduced based on the same approach towards hashing. Discriminative binary coding methods \cite{Rastegari:2012, He:2015}, like Random Maximum Margin Hashing approach, try to improve generalization by using SVMs to preserve the semantic structure in the Hamming space. Lin \emph{et al}. \cite{Lin:2014fs} proposed a supervised hashing method called FastHash that tries to capture the non-linearity in the feature space utilizing decision trees and uses a GraphCut based method for optimization. Supervised discrete hashing (SDH) \cite{Shen:2015fs} employs a linear support vector machine (SVM) to capture the similarity in the semantic space.

Despite the fact that new images are added to online photo databases every day, to the best of our knowledge, no supervised hashing method learns hash functions incrementally for newly added images. Efficiently updating hash functions with new images is a challenging task. Besides, some of the new images will possibly not belong to existing classes \emph{i.e.} forming new semantic classes, that makes this task even more challenging. For the existing hashing methods, such modifications to a database require recomputation of hash functions from scratch whenever a change occurs in a dynamic dataset, which is computationally intractable.

In this paper, we propose an incremental supervised hashing method with kernels based on binary and multi-class SVMs, which we refer to as \emph{Supervised Incremental Hashing} (SIH). We adopt the incremental learning fashion of SVMs in a hashing framework. Consequently, SIH can be easily extended to unseen classes incrementally. We identified three main objectives for our supervised hashing method -- as being incremental and parallelizable, avoiding overfitting by better generalization, and balancing ${+1}/{-1}$ in learned binary codes. The significance of balanced binary codes has been studied in the hashing literature \cite{Weiss:2009, Kulis:2009kr, liu2011hashing, he2011compact} and their absence leads to ineffective codes, especially when the code length is small and/or the sizes of semantic classes are unbalanced. To overcome these issues, we reformulate the supervised hashing task as a two-stage classification framework. In the first stage of classification, we use a binary SVM for each bit. The feature space is the input space of SVMs while binary codes are considered as class labels. In the second stage, the binary codes become the input space of a single multi-class SVM, and the semantic space is the output space of the SVM. Also, we penalize imbalance in binary codes in our optimization formulation.

We define an incremental strategy to learn SVMs and a \emph{discrete cyclic coordinate descent} (DCC) algorithm, similar to the one of SDH, to learn binary codes bit by bit as an approximate solution to the NP-hard discrete optimization problem. Our contributions can be summarized as follows:
\begin{itemize}
\item We propose a supervised hashing approach that provides better generalization with regularizations and maximizes the entropy by balancing binary codes. We formulate our hashing objectives in a single optimization task. 
\item We describe an algorithm that solves the optimization problem efficiently by an incremental strategy for training SVMs and an approximation to the solution of an NP-hard problem.
\item We define an incremental strategy for the proposed hashing method that takes the earlier hash functions and the final state of the database as its input and efficiently computes new hash functions which perform similarly to those computed from scratch on the final state of the database.
\end{itemize}

The rest of the paper is organized as follows: Section \ref{sec:method} introduces the proposed hashing method. Section \ref{sec:experiments} gives performance evaluation and comparison with the state-of-the-art supervised hashing methods, and Section \ref{sec:conclusion} provides conclusions.

\section{Incremental Hashing}
\label{sec:method}

\subsection{Problem Definition}

Given a set of data points $\{\mathbf{x}_i\}_{i=1}^n$, each $\mathbf{x}_i\in\mathbb{R}^d$, the goal of supervised hashing is to learn a function $H$ that maps data points to $m$-dimensional binary codes capturing underlying semantics. Several hashing methods \cite{Gionis:1999wf, Gong:2013kp} construct such a hash function by collecting a set of $m$ linear binary embeddings $H(\mathbf{x})=[h_1(\mathbf{x}),\ldots,h_m(\mathbf{x})]^\top$ such that $h_j(\mathbf{x})=\mathrm{sign}(\mathbf{w}_j^\top \mathbf{x}+\beta_j)$ where $\mathbf{w}_j\in\mathbb{R}^d$ is the normal vector and $\beta_j\in\mathbb{R}$ is the intercept of a hyperplane for $j=1,\ldots,m$. Some methods \cite{Kulis:2009kr, Liu:2012ii} employ a nonlinear embedding algorithm with a kernel function $\varphi:\mathbb{R}^d\times\mathbb{R}^d\rightarrow\mathbb{R}$ such that each binary embedding has the form $h_j(\mathbf{x})=\mathrm{sign}(\mathbf{w}_j^\top \boldsymbol\varphi(\mathbf{x})+\beta_j)$ where $\boldsymbol\varphi(\mathbf{x})=[\varphi(\mathbf{x},\mathbf{a}_1),\ldots,\varphi(\mathbf{x},\mathbf{a}_r)]^\top$ and $\{\mathbf{a}_l\}_{l=1}^s$ is a set of anchor points randomly selected from training samples.

\subsection{Learning Hash Functions for Dynamic Databases}

Modifications to an image database consist of adding new images to the database and deleting images from the database. Considering a database with class information, there are four types of modifications:
\begin{multicols}{2}
\begin{enumerate}
    \item Adding new classes,
    \item Adding images to existing classes,
    \item Deleting existing classes,
    \item Deleting images from existing classes.
\end{enumerate}
\end{multicols}
In this paper, we focus on the first three types of modifications. From a hashing perspective, the last type of modifications is the least interesting one because it technically results in reducing training size. Among the rest, deleting existing classes is the easiest case because we have already learned binary codes for all images in the final database. On the other hand, adding new classes is the hardest case since we have no prior information about the new classes.

We define three hashing strategies for dynamic datasets:
\begin{enumerate}
    \item \emph{Passive strategy}: Continuing to use the hash functions learned from an earlier state of the database \emph{i.e.} ignoring the changes in the training data.
    \item \emph{From-scratch strategy}: Learning hash functions on the final state of the database from scratch whenever a change occurs in the database.
    \item \emph{Incremental strategy}: Learning hash functions on the last state of the database incrementally starting from the previous hash functions.
\end{enumerate}
The incremental strategy can be considered as effective and efficient if its retrieval performance is similar to that of the from-scratch strategy while its training time is shorter. We first describe our supervised hashing method; then provide an incremental hashing strategy for our method in the following sections.

\subsection{Supervised Incremental Hashing}

We define a joint optimization task as a combination of $m$ binary linear SVMs and a multi-class linear SVM by Crammer and Singer \cite{Crammer:2002} and an imbalance penalty. Our method searches for optimal set of binary codes $\{\mathbf{b}_i\}_{i=1}^n$ where $\mathbf{b}_i=[b_{i1},\ldots,b_{im}]^\top$ corresponds to the $i$th data point in the training set. Let $\mathcal{Y}$ denote the set of ground truth semantic classes and $y_i\in\mathcal{Y}$ denote the class label of the $i$th data point; then we define a multi-class SVM model that maps binary codes to semantic classes \emph{i.e.} $\mathbf{b}_i\mapsto y_i$ for $i=1,\ldots,n$.

Let $\mathbf{B}\in\{-1,+1\}^{n\times m}$ be a matrix that collects binary codes where the $i$th row of $\mathbf{B}$ is equal to $\mathbf{b}_i^\top$. We define a binary SVM model treating $\{\boldsymbol\varphi(\mathbf{x}_i)\}_{i=1}^n$ as input data and column $j$ of $\mathbf{B}$ as binary class labels for $j=1,\ldots,m$. In addition, we add a new term that penalizes imbalanced assignments of ${+1}/{-1}$ in binary codes. Our formulation for supervised hashing is given by
\begin{equation}
\label{eq:svmhash}
\begin{split}
    \min_{\mathbf{B},\mathcal{W},\Xi} &\quad\lambda\Biggl(\frac{1}{2}\sum_{k\in\mathcal{Y}}\|\mathbf{w}_k^b\|^2 + C^b\sum_{i=1}^n\xi^b_i\Biggr)
    +\sum_{j=1}^m \Biggl(\frac{1}{2}\|\mathbf{w}_j^x\|^2 + C^x\sum_{i=1}^n\xi^x_{ij}\Biggr) + \gamma\sum_{j=1}^m \biggl|\sum_{i=1}^n b_{ij}\biggr|\\
    \mathrm{s.t.} & \quad \forall (i,k)\quad(\mathbf{w}^b_{y_i}-\mathbf{w}^b_k)^\top\mathbf{b}_i\geq \mathbbm{1}[y_i\neq k]-\xi_i^b,\\
    & \quad \forall (i,j)\quad b_{ij}\,(\mathbf{w}_{j}^x)^\top\boldsymbol\varphi(\mathbf{x}_i)\geq1-\xi_{ij}^x,\\
    &\quad \forall (i,j)\quad \xi_{ij}^x\geq0,\\
    &\quad\mathbf{b}_i\in\{-1,+1\}^m
\end{split}
\end{equation}
where $\lambda$ and $\gamma$ are scaling parameters, $C^x$ and $C^b$ are soft margin parameters of the SVMs. Note that the sum of column $j$ in $\mathbf{B}$ is equal to zero when bit $j$ is balanced. There will be no penalty for bit $j$ in this case. This imbalance penalty is also important for the performance of the binary SVMs with the fixed parameter $C^x$ since they are sensitive to imbalance in the datasets \cite{Batuwita:2013}. We can add an extra element 1 to the vectors of all data points and binary codes in order to add a bias term to the loss function of SVMs. Finally, our hash function is defined as $H(\mathbf{x}) = \mathrm{sign}\bigl((\mathbf{W}^x)^\top\boldsymbol\varphi(\mathbf{x})\bigr)$ where the $j$th column of $\mathbf{W}^x\in\mathbb{R}^{d\times m}$ is equal to $\mathbf{w}_j^x$ for $j=1,\ldots,m$. Similarly, we define another weight matrix $\mathbf{W}^b\in\mathbb{R}^{d\times m}$ such that its $j$th column is equal to $\mathbf{w}_k^b$ for $k=1,\ldots,|\mathcal{Y}|$. This optimization task can be solved by an alternating algorithm that consists of two main steps: Training SVMs and learning binary codes as follows.

\subsection{Training SVMs}

If we fix all binary codes $\mathbf{B}$, the optimization problem \eqref{eq:svmhash} will be reduced to $m$ separate binary SVM and one multi-class linear SVM problems. These $(m+1)$ SVMs can be trained in parallel since they are conditionally independent on binary codes. We employ an efficient large-scale SVM learning technique called the \emph{Optimized Cutting Plane Algorithm for SVMs} (OCAS) \cite{Franc:2009}. The OCAS algorithm is based on approximating the original convex loss function of an SVM by a piecewise linear function defined as the maximum over a set of linear under-estimators called cutting planes. At each step, a new cutting plane is added to the set with a cost of $\mathcal{O}(nd)$ complexity. Each under-estimator is a linear approximation to the loss function of the SVM at a weight point $\mathbf{w}'$. The method also has a multi-class counterpart called OCAM.

\begin{figure*}[!ht]
\centering
\includegraphics[width=.775\linewidth]{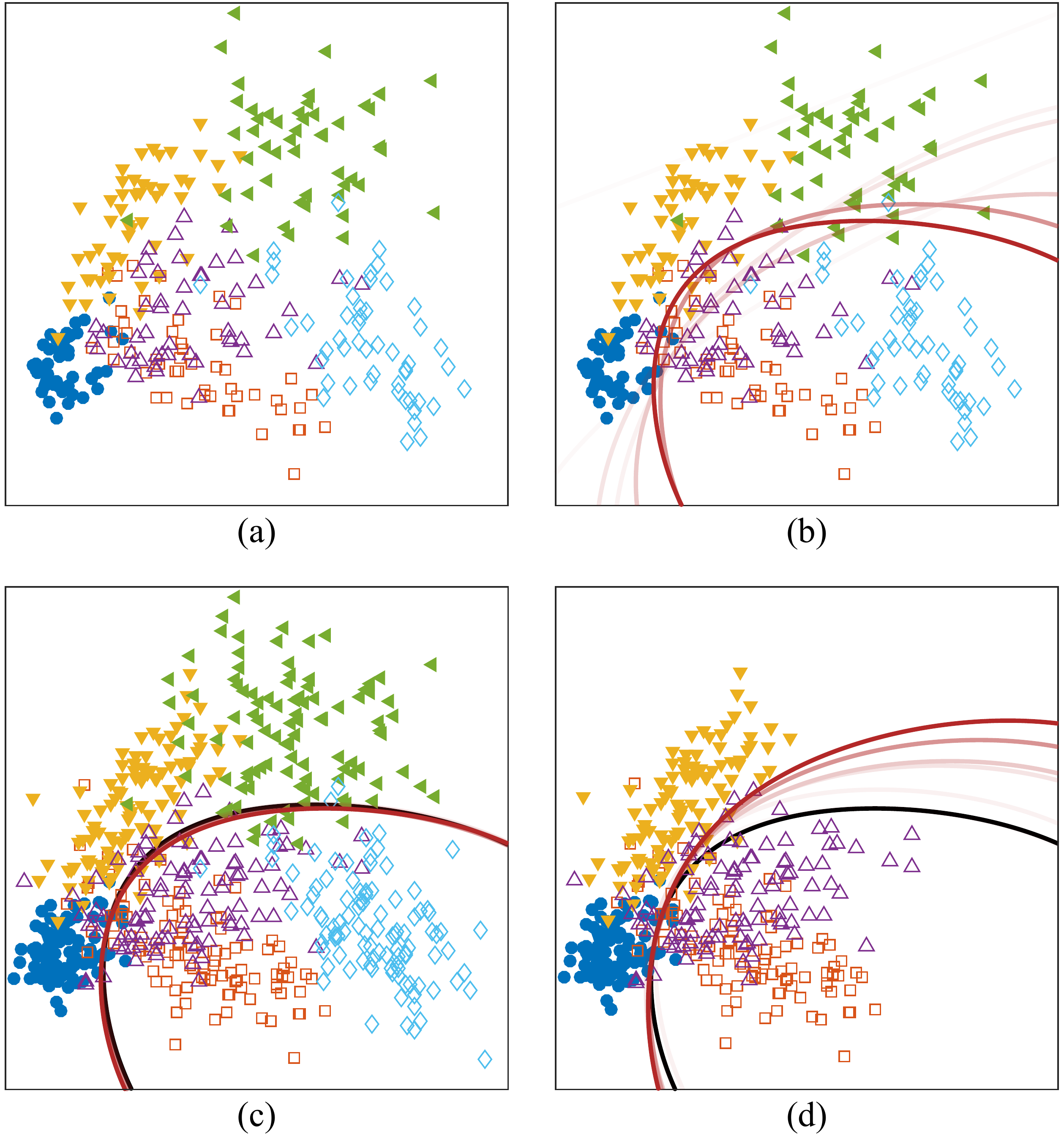}
\caption{(a) A sample dataset of 300 data points from 6 classes represented by colors and shapes. Assignments of $+1$ and $-1$ are indicated by filled and empty shapes, respectively. (b) Each red line represents a hyperplane corresponding to the weight vector at each step of the OCAS method \emph{i.e.} $(\mathbf{w}^b_t)^\top\mathbf{x}=0$. Increasing transparency of the lines indicates earlier iterations of the execution \emph{i.e.} smaller~$t$. (c) The number of data points is increased to 600 by adding new points from the same distributions in (b). Red lines represent the hyperplanes when the OCAS method is initialized with the solution at (b) shown by a black line. (d) Two classes are deleted from (c). Red lines represent the hyperplanes when the OCAS method is initialized with the solution at (c) shown by a black line.}
\label{fig:inc_SVM}
\end{figure*}

The OCAS method tries to minimize the total number of iterations by additional computations at each step. Let $F(\mathbf{w})=\tfrac{1}{2}\|\mathbf{w}\|^2+C\,R(\mathbf{w})$ be an SVM objective function such that $\|\partial R(\mathbf{w})\|\leq G$, then the OCAS method converges after at most $$\log_2\frac{F(\mathbf{0})}{4C^2G^2}+\frac{8C^2G^2}{\varepsilon}-2$$ iterations where $\varepsilon$ denotes a tolerance value. We can further reduce the total number iterations for an SVM with hinge loss by $\log_2\tfrac{nC}{F(\mathbf{w}_0)}$ if there exists an initial weight vector $\mathbf{w}_0$ better than the zero vector \emph{i.e.} $F(\mathbf{w}_0)<F(\mathbf{0})$ (See Appendix). Considering the cases of adding new images to existing classes and deleting classes, the solution to the previous SVM problem can be a good starting point. In addition, changes in binary codes $\mathbf{B}$ from one iteration to the next one are usually small during the execution of our algorithm. Therefore, we take advantage of the pre-computed SVMs from the previous iteration in an incremental approach. We employ a warm start strategy for training SVMs similar to \cite{Tsai:2014}. We initialize the \emph{best-so-far solution} in the OCAS algorithm with the solution to the SVM in the previous iteration instead of a zero vector for each bit. This initialization results in a reduced number of iterations in the OCAS computation and fewer cutting planes. If there exists no change in column $j$ of $\mathbf{B}$ between iteration $(t-1)$ and iteration $t$, the corresponding SVM is not trained again. Similarly, the same approach is used for the multi-class linear SVM using the OCAM algorithm. Our hashing method converges when the entire matrix $\mathbf{B}$ remains unchanged. A simulation of our incremental SVM approach is presented on a sample dataset in Figure~\ref{fig:inc_SVM}.

\subsection{Learning Binary Codes}

In the process of learning binary codes $\mathbf{B}$, we fix all the SVM weights $\{\mathbf{w}^x_j\}_{j=1}^m$ and $\{\mathbf{w}_k^b\}_{k\in\mathcal{Y}}$ in \eqref{eq:svmhash}. This results in an integer programming problem and is therefore NP-hard. As a result, we take an approach akin to the discrete cyclic coordinate descent algorithm in \cite{Shen:2015fs}. We learn binary codes column by column in $\mathbf{B}$ iteratively based on the rest of the binary codes until convergence. For updating bit $j$ in $\mathbf{b}_i$ for $i=1,\ldots,n$, we reduce \eqref{eq:svmhash} to the following optimization problem:
\begin{equation}
\begin{split}
    \min_{\{b_{ij}\}_{i=1}^n} &\quad \sum_{i=1}^n L(b_{ij},i,j) + \gamma\biggl|\sum_{i=1}^n b_{ij}\biggr|\\
    \textrm{s.t.}&\quad \forall i\quad b_{ij}\in\{-1,+1\}
\end{split}
\end{equation}
where $L(z,i,j)$ represents the total hinge losses depending specifically on data point $i$ when the corresponding bit $j$ is equal to $z\in\{-1,+1\}$, specifically:
\begin{equation*}
\begin{split}
    L(z,i,j) &= \beta\, C^b\max_{k\in\mathcal{Y}}\bigl(\mathbbm{1}[y_i\neq k]+z\,(w_{k,j}^b-w_{y_i,j}^b)+\theta_{ijk}\bigr)
    + C^x\max\bigl(0, 1-z\,(\mathbf{w}_{j}^x)^\top\boldsymbol\varphi(\mathbf{x}_i)\bigr)\\
\end{split}
\end{equation*}
where $\theta_{ijk}=\sum_{u\neq j} b_{iu}(w_{k,u}^b-w_{y_i,u}^b)$ is a bias term that depends on the binary codes excluding bit $j$. In case of no penalty for imbalance ($\gamma=0$), the closed-form solution is $b_{ij} =\argmin_{z\in\{-1,+1\}} L(z,i,j)$ for $i=1,\ldots,n$. In our case, we need to consider the trade-off between the hinge losses and the imbalance penalty. This is achieved by sorting the data points with respect to the difference $\delta_{ij}=L(-1,i,j)-L(+1,i,j)$ and finding a cutting location between ${+1}/{-1}$ assignments as follows. Let $I$ contain the indices of data points in ascending order, then the cutting location is determined by
\begin{equation}
\label{eq:cut}
    \mathsf{cut}=\argmin_{l\in\{0,\ldots,n\}}\biggl(\gamma|2l-n|+\sum_{i=1}^{l}L\bigl(-1,I[i],j\bigr)
    +\sum_{i=l+1}^nL\bigl(+1,I[i],j\bigr)\biggr).    
\end{equation}
Next, we assign $-1$ to bit $j$ of data points from $I[1]$ to $I[\mathsf{cut}]$ and $+1$ for data points from $I[\mathsf{cut}+1]$ to $I[n]$. Note that the first sum in \eqref{eq:cut} represents the cumulative hinge losses for the assignments of $-1$ and the second one represents the same for $+1$ assignments. We repeat these column updates for $j=1,\ldots,m$ until $\mathbf{B}$ converges; that typically requires a few full updates of $\mathbf{B}$. The proposed Supervised Incremental Hashing (SIH) method is summarized in Algorithm~\ref{alg:svmHash}.

\begin{algorithm}[!ht]
\caption{Supervised Incremental Hashing}
\label{alg:svmHash}
\begin{algorithmic}[1]
    \REQUIRE Training data $\{(\mathbf{x}_i,y_i)\}_{i=1}^n$, code length $m$, number of anchor points $r$, maximum iteration number $\mathsf{max\_iter}$, parameters $C^x$, $C^b$, $\lambda$ and $\gamma$.
    \ENSURE Binary codes $\mathbf{B}$, hash function $H(\mathbf{x}) = \mathrm{sign}\bigl((\mathbf{W}^x)^\top\mathbf{x}\bigr)$.
    \vspace{6pt}
    \STATE Randomly select $r$ anchor points $\{\mathbf{a}_i\}_{i=1}^s$ from the training data 
    \STATE Compute the kernel function $\boldsymbol\varphi(\mathbf{x}_i)$ for $i=1,\ldots,n$.
    \STATE Initialize binary codes $\mathbf{B}^{(0)}$ with a random string in $\{+1,-1\}^{m}$ for each class.
    \FOR{$t\leftarrow 1$ to $\mathsf{max\_iter}$ }
    \FOR{$j\leftarrow1$ to $m$ \textbf{in parallel}}
    \IF {column $j$ of $\mathbf{B}^{(t-1)}\neq$ column $j$ of $\mathbf{B}^{(t)}$,}
    \STATE Train binary SVM for bit $j$ using the OCAS algorithm on $\bigl\{\bigl(\boldsymbol\varphi(\mathbf{x}_i),b_{ij}^{(t-1)}\bigr)\bigr\}_{i=1}^n$.
    \ENDIF
    \ENDFOR
    \STATE Train linear multiclass SVM using the OCAM algorithm where the data is $\bigl\{\bigl(\mathbf{b}^{(t-1)}_i,y_i\bigr)\bigr\}_{i=1}^n$.
    \REPEAT
    \FOR{$j$ from $1$ to $m$}
    \STATE Compute $\mathsf{cut}$ in (3) of the main paper
    \STATE Update column $j$ in $\mathbf{B}^{(t)}$ according to $\mathsf{cut}$ and $I$. 
    \ENDFOR
    \UNTIL{$\mathbf{B}^{(t)}$ convergences}
    \IF {$\mathbf{B}^{(t-1)}=\mathbf{B}^{(t)}$}
    \STATE \textbf{break} the loop 
    \ENDIF
    \ENDFOR
\end{algorithmic}
\end{algorithm}

\subsection{Incremental Updates to Hash Functions}

Our method can be efficiently updated whenever a modification occurs in the training set. We describe the initialization of our incremental learning strategy from the easiest to the hardest type of modification as follows:
\begin{enumerate}
    \item \emph{Deleting existing classes}: Let $\mathcal{Y}_d$ be the set of deleted class labels, we remove $\mathbf{w}^y_k$ for $k\in\mathcal{Y}_d$ from the multi-class SVM $\{\mathbf{w}_k^b\}_{k\in\mathcal{Y}}$ and the rows of $\mathbf{B}$ that correspond to images associated with any class in $\mathcal{Y}_d$ and start our incremental strategy to adopt to the changes.
    \item \emph{Adding images to existing classes}: For each class, the most frequent binary string pattern from $\mathbf{B}$ is found. Next, each new image is initialized with the corresponding binary code according to its class and the hashing functions are changed incrementally with our method.
    \item \emph{Adding new classes}: New rows to $\mathbf{B}$ are added corresponding to new images, and we initialize each new class with a random binary string. Since we do not have multi-class SVM trained on the new classes, we only train the multi-class SVM $\{\mathbf{w}_k^b\}_{k\in\mathcal{Y}_f}$ on $\mathbf{B}_f$ from scratch where the subscript $f$ denotes the final version. The hashing functions are learned with our incremental algorithm.    
\end{enumerate}

\section{Experiments}
\label{sec:experiments}

We conducted extensive experiments to assess the effectiveness and efficiency of the proposed method, SIH. Before evaluating the performance on dynamic datasets, we first compared our method with the state-of-the-art supervised hashing techniques including CCA-ITQ \cite{Gong:2013kp}, KSH \cite{Liu:2012ii}, FastHash \cite{Lin:2014fs}, and SDH \cite{Shen:2015fs}. The performance of the methods was analyzed regarding search accuracy and training/testing time on three large-scale datasets -- CIFAR-10, MNIST, and NUS-WIDE. All experiments were performed in the MATLAB environment on a machine with a 2.8 GHz Intel Core i7 CPU and 16GB RAM using the public code provided by the authors with their suggested parameters unless otherwise specified.

\subsection{Datasets and Experimental Setup}

The CIFAR-10 dataset \cite{Krizhevsky:2009ak} has 60,000 labeled images from 10 classes of vehicles and animals. There exist 50,000 training and 10,000 test images in the dataset. In our experiments, each image is represented by a GIST descriptor \cite{Oliva:2001at} of 512 dimensions. The MNIST dataset \cite{Lecun:1998yb} is a collection of 28$\times2$8 pixel images of handwritten digits. The dataset contains a training set of 60,000 examples and a test set of 10,000 examples. The NUS-WIDE dataset \cite{Chua:2009tj} includes 269,648 images and associated semantic labels of 81 concepts from Flickr. Unlike the other two datasets, each image is associated with zero or more labels. Therefore, we only used the training images that are associated with exactly one of the most frequent 10 tags. The resultant training set has 38,255 images with 10 classes. Similarly, we constructed a query set of 25,386 images among the test images associated with only one of those tags. A 500-dimensional bag-of-words vector is used to represent each image in our experiments where the codebook is generated from SIFT descriptors.

For the KSH method, 5,000 images are sampled uniformly from the training part of each dataset because of long training time of this technique (Table~\ref{table:timing}). The tree depth parameter of FastHash is set to 2 due to its higher computational complexity. We sample 1,000 images uniformly from the training sets as anchor points $\{\mathbf{a}_l\}_{l=1}^s$ for all the hashing methods with kernels (SIH, KSH, and SDH). All these kernel methods used an RBF kernel $\varphi(\mathbf{x},\mathbf{a})=\exp\bigl(-\|\mathbf{x}-\mathbf{a}\|^2/2\sigma^2\bigr)$ with a kernel width $\sigma$ which is adjusted for each dataset using cross-validation. For all datasets, features are first centered to zero; then each data sample is normalized to have the unit length.

For SIH, we adjusted the soft margin parameters of the SVMs and the kernel width based on cross-validation on some binary codes obtained from other hashing methods. Next, we empirically adjusted the other parameters $\lambda$ and $\gamma$ for each dataset. We prefer the multi-class SVM dominate the process of learning binary codes by setting $\lambda$ to a large number. For example, for the CIFAR-10 dataset the parameter values are $C^x=16$, $C^b=10^{-3}$, $\lambda=m\times 10^8$ and $\gamma=10^5$. The maximum number of iterations is set to 5 like SDH.

\subsection{Evaluation Methodology}

We employed leave-one-out validation on test data in our experiments to assess the quality of hash functions learned from training data. We used each image in the test data as a query while we treat the remaining test images as a retrieval set. For each query, we rank images according to Hamming distances between their binary codes and that of the given query. For quantitative analysis, we compute precision and recall value at each Hamming radius. Next, these values are used for computing mean average precision (mAP) and also mean precision values.

\subsection{Effects of Training and Anchor Set Size}

We start our experiments by analyzing the effects of training and anchor set size on retrieval. The SIH method is analyzed in details with different numbers of anchor points and training points for comparison (Table~\ref{table:timing}). As expected, larger training and anchor sets provide better performance with longer execution time. The number of anchor points has a greater influence on retrieval performance than training size. Note that the test time is affected by the scale of the anchor set. Our method outperforms other methods in retrieval performance while it has competitive execution time. Note that the training time in Table~\ref{table:timing} can be enhanced by a computer with a larger number of nodes as our method has a distributed inference algorithm.

\begin{table*}[!ht]
  \caption{Our method (SIH) is compared in terms of mean average precision (mAP), mean of precision at Hamming radius $r = 2$, training and test time for 32-bit hash codes on the CIFAR-10 dataset. The experiments were performed on a machine with an Intel quad-core processor.}
  \label{table:timing}
\begin{center}
  \begin{tabular}{|l||r|r|r|r|r|r|}
  \hline
   \multicolumn{1}{|c||}{\multirow{2}{*}{\bfseries Method}} & \multicolumn{1}{c|}{\bfseries Training} & \multicolumn{1}{c|}{\bfseries Anchor} &  \multicolumn{1}{c|}{\multirow{2}{*}{\bfseries mAP}} & \multicolumn{1}{c|}{\bfseries Precision} & \multicolumn{1}{c|}{\bfseries Training} & \multicolumn{1}{c|}{\bfseries Test}\\
   & \multicolumn{1}{c|}{\bfseries Set Size} & \multicolumn{1}{c|}{\bfseries Set Size} &  & \multicolumn{1}{c|}{\bfseries at $r=2$} & \multicolumn{1}{c|}{{\bfseries Time} (s)} & \multicolumn{1}{c|}{{\bfseries Time} ($\mu$s)}\\
    \hline
    \hline
    \multicolumn{1}{|l||}{\multirow{6}{*}{SIH}}& 5,000 & 300 & 0.322 & 0.383 & 2.7 & 6.1 \\
    & 5,000 & 1,000 & 0.366 & 0.420 & 7.5 & 15.9 \\
    & 5,000 & 3,000 & 0.393 & 0.438 & 47.3 & 45.7 \\
    & 50,000 & 300 & 0.360 & 0.417 & 61.7 & 6.3 \\
    & 50,000 & 1,000 & \textbf{0.429} & 0.478 & 171.1 & 17.7 \\
    \hline
    SDH & 50,000 & 1,000 & 0.422 & \textbf{0.478} & 25.0 & 24.9 \\
    \hline
    Fasthash & 50,000 & \multicolumn{1}{|c|}{\quad-} & 0.357 & 0.271 & 628.4 & 102.6 \\
    \hline    
    KSH & 5,000 & 1,000 & 0.343 & 0.322 & 2,802.8 & 26.8\\   
    \hline                     
    CCA-ITQ & 50,000 & \multicolumn{1}{|c|}{\quad-} & 0.325 & 0.410 & 1.5 & 3.8 \\
    \hline
  \end{tabular}
\end{center}
\end{table*}

\subsection{Retrieval Performance Analysis}

\begin{figure*}[!ht]
\centering
\includegraphics[width=.975\linewidth]{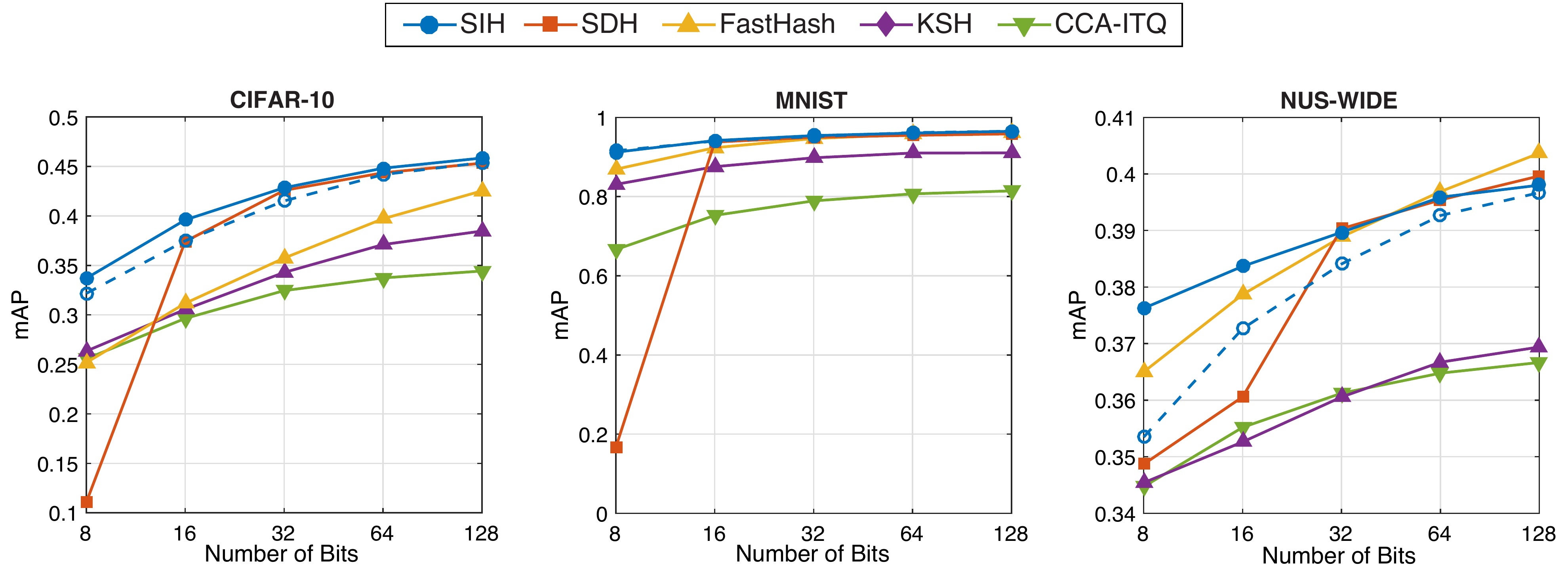}
\caption{Our method (SIH) is compared with the state-of-the-art methods on, from left to right, the CIFAR-10, MNIST and NUS-WIDE datasets in terms of mean average precision (mAP). Dashed line represents SIH without imbalance penalty.}
\label{fig:retrieval} 
\end{figure*}

\begin{figure*}[ht]
\centering
\includegraphics[width=.975\linewidth]{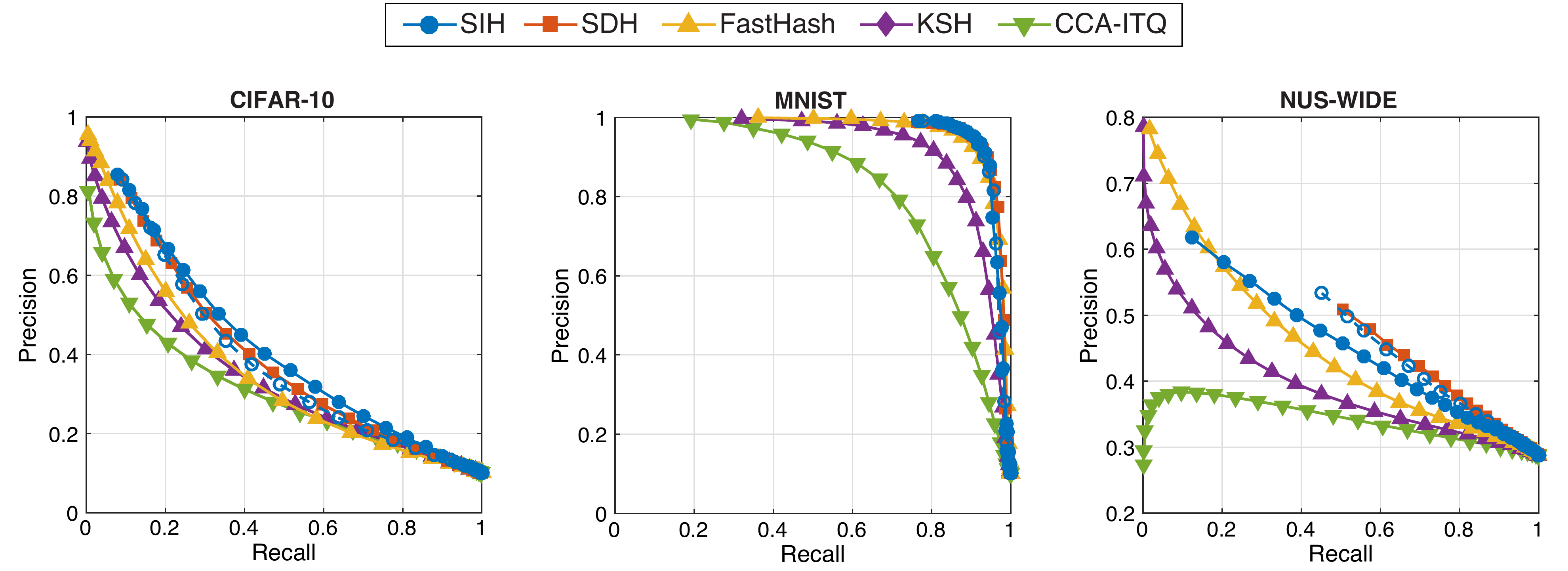}
\caption{Our method (SIH) is compared with the state-of-the-art methods on, from left to right, the CIFAR-10, MNIST and NUS-WIDE datasets by precision-recall curves for 32-bit length hash codes. Dashed line represents SIH without imbalance penalty.}
\label{fig:recprec}
\end{figure*}

We report retrieval performance as mean average precision (mAP) for all methods on the three datasets in Figure~\ref{fig:retrieval} for different code lengths. Precision-recall curves of all methods on all datasets are displayed in Figure~\ref{fig:recprec} for 32-bit length codes. As seen in the table and figures, the proposed hashing method outperforms the state-of-the-art supervised hashing methods on the CIFAR-10 and MNIST datasets, in which each image belongs to a single class. In addition, the imbalance penalty improves the retrieval performance for almost all cases. Note that we used a sequential sampling procedure for initializing $\mathbf{B}$ where balanced binary codes have a higher probability for the SIH version with imbalance penalty \emph{i.e.} $\gamma>0$. SDH provides the second best performance. On the other hand, our method has the best performance on the NUS-WIDE dataset for shorter binary codes, and it has competitive results along with FastHash and SDH for longer codes.

\subsection{Retrieval Performance Analysis for Dynamic Datasets}

\begin{figure*}[!ht]
\centering
\includegraphics[width=.975\linewidth]{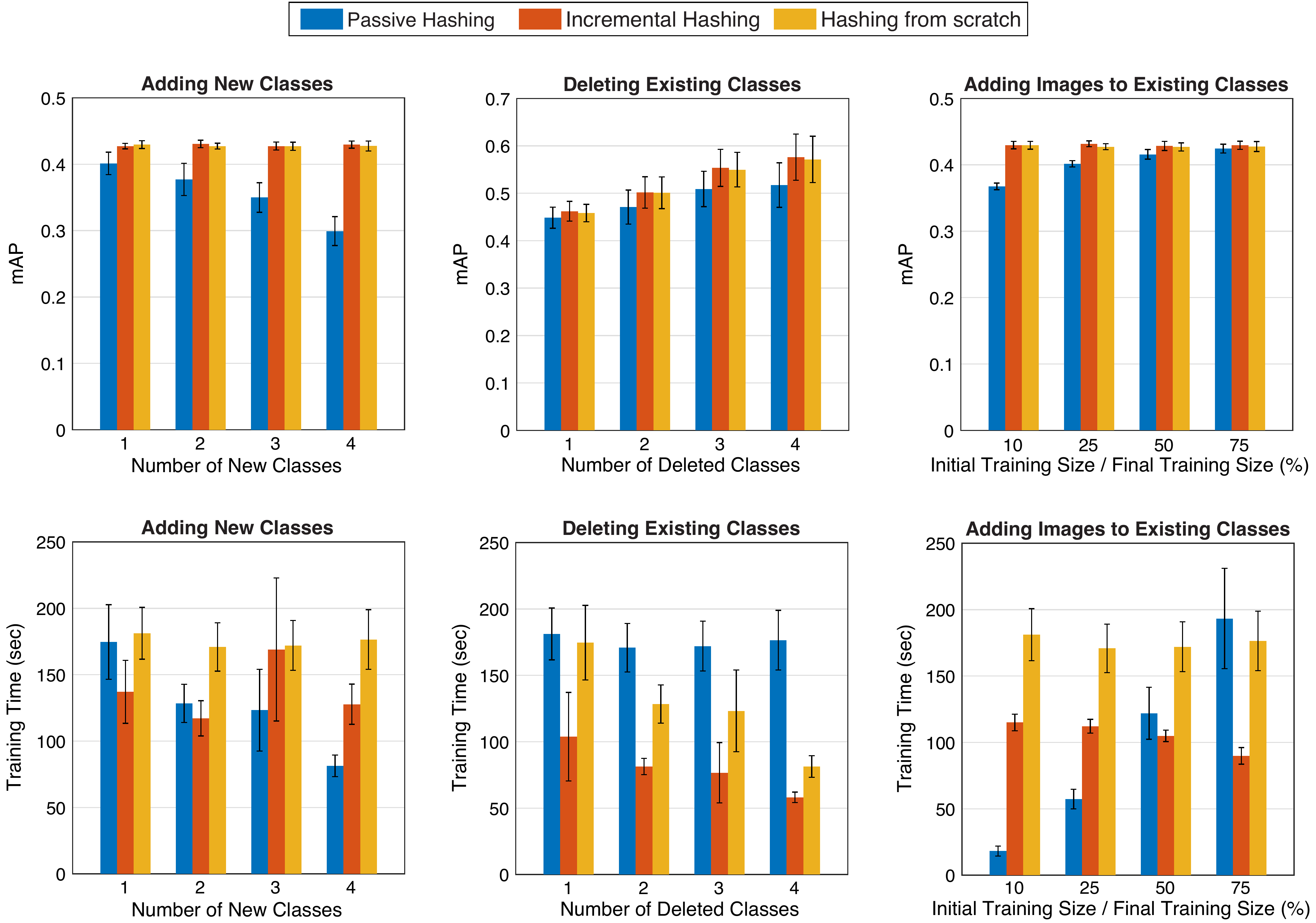}
\caption{Incremental Hashing is compared with from-scratch and passive hashing for different types of modifications, from left to right, adding new classes, deleting existing classes and adding new images to existing classes on CIFAR-10 in terms of mean average precision (mAP) in the first row and in terms of training time in the second row at 32-bits.}
\label{fig:inc}
\end{figure*}

\begin{figure*}[!ht]
\centering
\includegraphics[width=.975\linewidth]{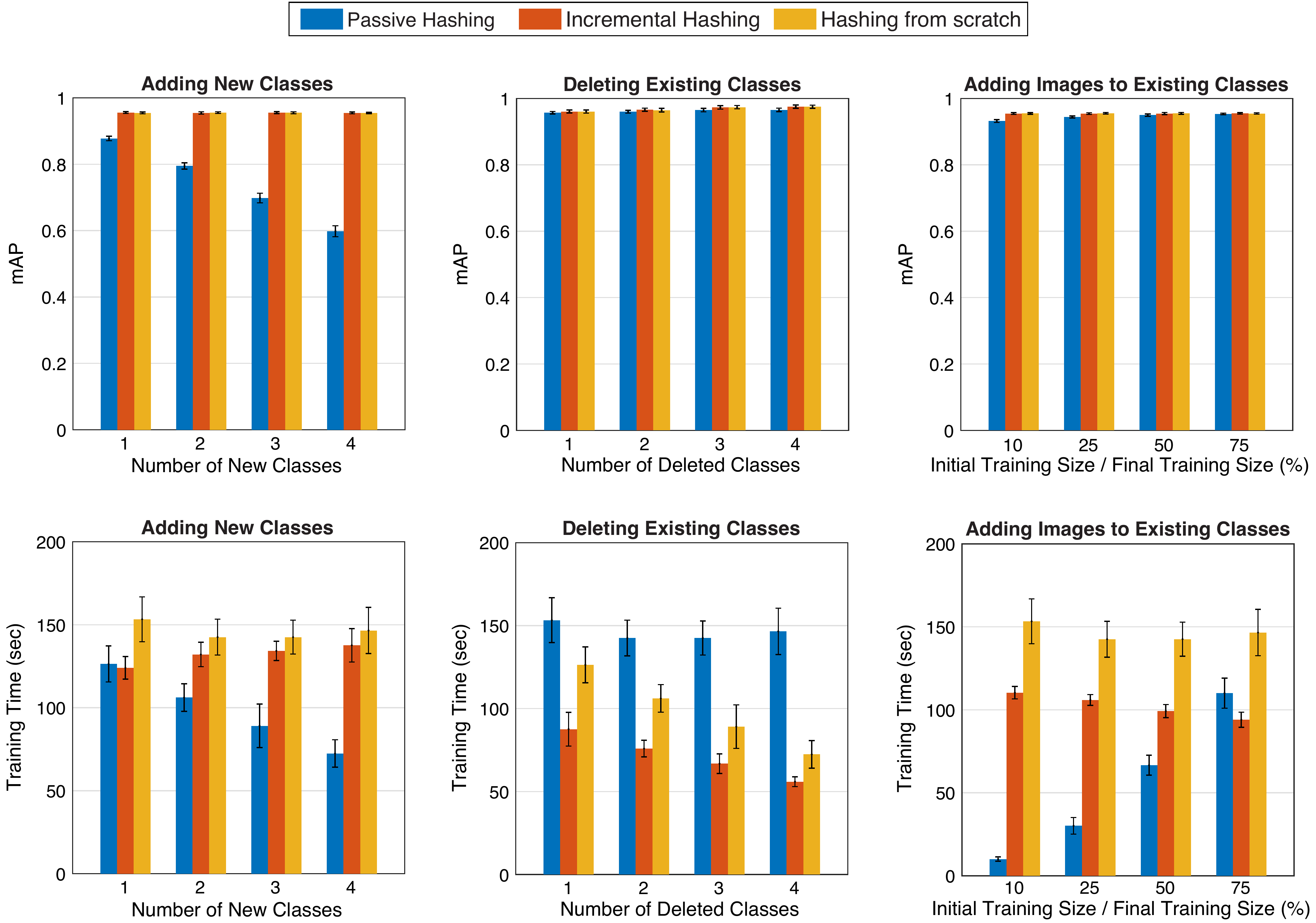}
\caption{Incremental Hashing is compared with from-scratch and passive hashing for different types of modifications, from left to right, adding new classes, deleting existing classes and adding new images to existing classes on MNIST in terms of mean average precision (mAP) in the first row and in terms of training time in the second row at 32-bits.}
\label{fig:inc_mnist}
\end{figure*}

\begin{figure*}[!ht]
\centering
\includegraphics[width=.975\linewidth]{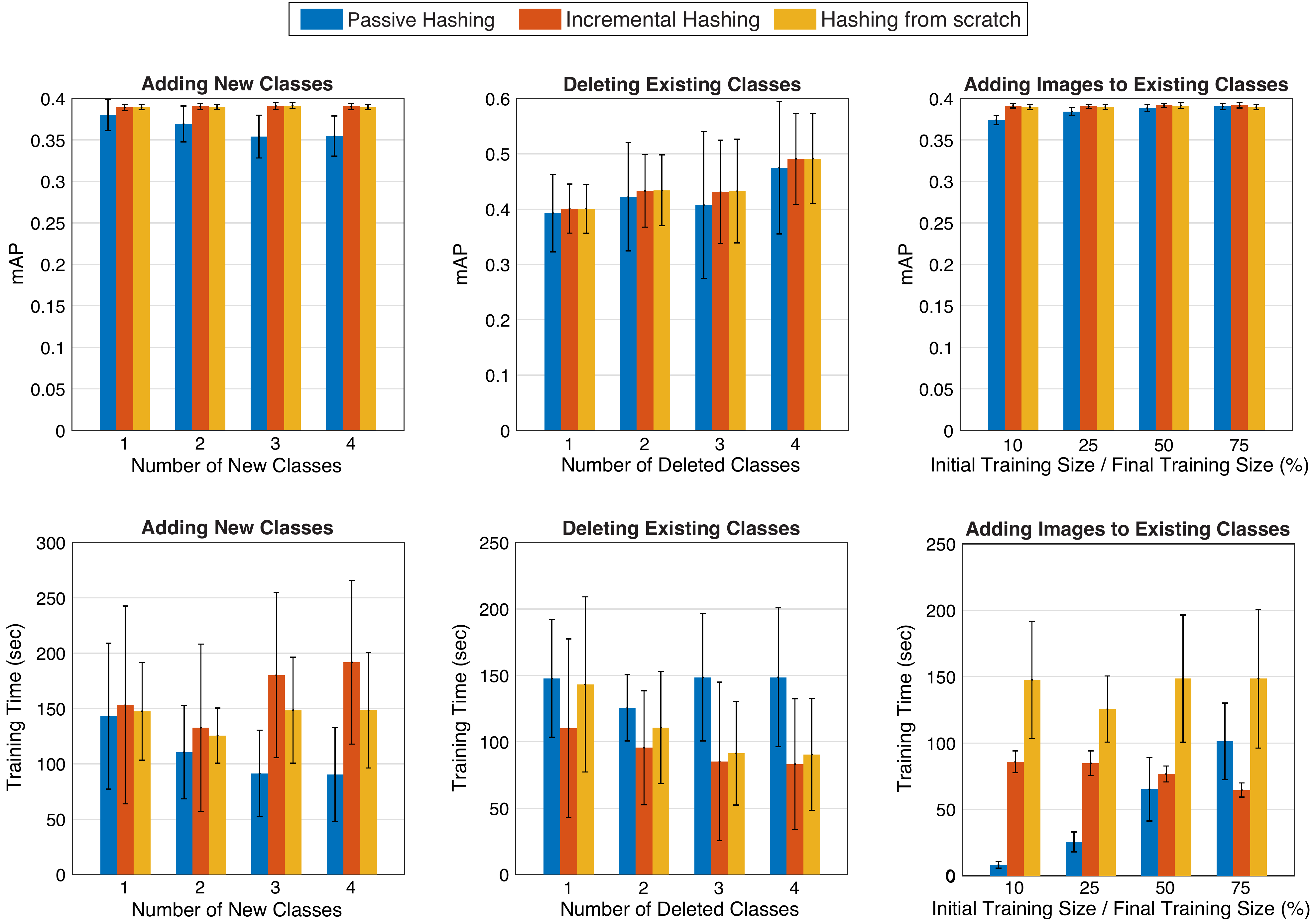}
\caption{Incremental Hashing is compared with from-scratch and passive hashing for different types of modifications, from left to right, adding new classes, deleting existing classes and adding new images to existing classes on NUS-WIDE in terms of mean average precision (mAP) in the first row and in terms of training time in the second row at 32-bits.}
\label{fig:inc_nuswide}
\end{figure*}

We evaluated our incremental hashing strategy in comparison to the passive and from-scratch strategies for the aforementioned three types of modifications on the CIFAR-10 dataset with 32 bits binary codes in terms of mAP scores and training time. For the case of deleting classes, we first learned hash functions from the entire training data. Next, we removed images from randomly selected classes and learned hash functions both from scratch and incrementally on the final training data. We repeated this 25 times and reported mAP and training time. A similar methodology in the reverse order was employed for adding new classes case. For adding new images to existing classes case, we first learned hash functions from a randomly selected subset of training data and then used the entire training set for from-scratch and incremental strategies. For adding new classes and new images to existing classes, we used the same query set as in the previous section. For deleting classes case, we removed the images associated with those classes and computed mAP for the rest of the queries. The mAP and training time are shown in Figures~\ref{fig:inc}-\ref{fig:inc_nuswide} for three types of modifications on all datasets. Note that the training time reported for the passive strategy represents the initial computation time, and it does not have any additional computation for changes in the dataset. Our incremental hashing strategy reaches the same retrieval performance as the from-scratch strategy while requiring shorter training time. Note that the total computation time for both training on 10\% of training data and incrementally training on the entire data is less than the training time on the entire data set from scratch. As a result, it might be a sound strategy to use our supervised hashing method as a combination of training on a small set of representative images followed by incremental training on the entire training data. This will reduce the training time of our method in Table~\ref{table:timing}.

\section{Conclusion}
\label{sec:conclusion}

We presented an incremental supervised hashing method based on a two-stage classification framework. We formulated a joint optimization task for the classification problem and the problem of finding optimal binary codes. An efficient algorithm was developed for learning hash functions in a distributed scheme where the sub-problems are solved independently. Experiments validate that the incremental hashing strategy for dynamic datasets is capable of updating hash functions efficiently. Besides, the proposed approach provides higher quality codes with well-balanced bits and better generalization.

%\subsubsection*{Acknowledgments}

%This research is based upon work supported by the Office of the Director of National Intelligence (ODNI), Intelligence Advanced Research Projects Activity (IARPA), via IARPA R&D Contract No. 2014-14071600012. The views and conclusions contained herein are those of the authors and should not be interpreted as necessarily representing the official policies or endorsements, either expressed or implied, of the ODNI, IARPA, or the U.S. Government. The U.S. Government is authorized to reproduce and distribute reprints for Governmental purposes notwithstanding any copyright annotation thereon.

\begin{small}

\bibliography{svmHash_bmvc16}

\end{small}

\section*{Appendix}
\label{sec:app}

The SVM problem can be described as an unconstrained regularized risk minimization problem as follows:
\begin{equation}
	\mathbf{w}^*=\argmin_{\mathbf{w}\in\mathbb{R}^d} F(\mathbf{w}):=\frac{1}{2}\|\mathbf{w}\|^2+C\,R(\mathbf{w}),
\end{equation}
where $\mathbf{w}\in\mathbb{R}^d$ denotes the weight vector to be learned, $\tfrac{1}{2}\|\mathbf{w}\|^2$ is a quadratic regularization term, $C>0$ is a fixed regularization constant and $R:\mathbb{R}^d\rightarrow\mathbb{R}$ is a non-negative convex risk function \emph{e.g.} hinge loss on training data, $R(\mathbf{w})=\sum_{i=1}^n\max(0,1-y_i\mathbf{w}^\top\mathbf{x}_i)$.

The cutting plane method approximates the convex function $F(\mathbf{w})$ by a piecewise linear function $F_t(\mathbf{w})$ with $t$ cutting planes. Let $\mathbf{w}_t$ and $\mathbf{w}_t^b$ be the \emph{current solution} and the \emph{best-so-far solution} of the OCAS method at iteration $t$, respectively. The stopping condition of the OCAS method is defined for a given tolerance parameter $\varepsilon$ as $F(\mathbf{w}_t^b)-F_t(\mathbf{w}_t)\leq\varepsilon$. Given in \cite{Franc:2009}, assume that $\|\partial R(\mathbf{w})\|\leq G$ for all $\mathbf{w}\in\mathbb{R}^d$, and $F(\mathbf{w}_t^b)-F_t(\mathbf{w}_t)=\varepsilon_t>0$, then
\begin{equation}
	\varepsilon_t-\varepsilon_{t+1}\geq \frac{\varepsilon_t}{2}\min\biggl(1, \,\frac{\varepsilon_t}{4C^2G^2}\biggr).
	\label{eq:et}
\end{equation}
From \eqref{eq:et}, for any $\varepsilon_t>4C^2G^2$, we say that $\varepsilon_{t+1}\leq \tfrac{\varepsilon_t}{2}$; starting from a weight vector $\mathbf{w}_0$, we can reach at a level precision better than $4C^2G^2$ after at most $\log_2\tfrac{F(\mathbf{w}_0)}{4C^2G^2}$ iterations \cite{Franc:2009}. Subsequently, we can find the remaining number of iterations by solving the following differential equation:
\begin{equation}
	\varepsilon_{t+1}+\frac{\varepsilon_t^2}{8C^2G^2}-\varepsilon_t=0.
\end{equation}
Franc and Sonnenburg provide a solution to this equation in \cite{Franc:2009}. We need $\tfrac{8C^2G^2}{\varepsilon}-2$ more iterations until convergence. The OCAS method is initialized its computation from the zero vector. Therefore, the total number of iterations of the OCAS method to solve an SVM problem is at most
\begin{equation}
	\log_2\frac{F(\mathbf{0})}{4C^2G^2}+\frac{8C^2G^2}{\varepsilon}-2.
\end{equation}

In Supervised Incremental Hashing, we use SVMs with hinge losses. Thus, $F(\mathbf{0})$ is equal to $n\,C$ for a dataset with $n$ images. The initial weight vector $\mathbf{w}_0$ has no effect on the number of iterations following a precision level better than $4C^2G^2$. However, considering an initial vector $\mathbf{w}_0$ in our incremental setting such that $F(\mathbf{w}_0)<n\,C$ \emph{i.e.} having a better solution $\mathbf{w}_0$ than the zero vector, we can reduce the number of iterations until achieving such a precision level by at most 
\begin{equation}
	\log_2\frac{F(\mathbf{0})}{4C^2G^2}-	\log_2\frac{F(\mathbf{w}_0)}{4C^2G^2}
	= \log_2\frac{n\,C}{F(\mathbf{w}_0)}.
\end{equation}
This concludes the claim in Section~\ref{sec:method}.

\end{document}